\documentclass[10pt, conference]{IEEEtran}
\IEEEoverridecommandlockouts

\usepackage{cite}
\usepackage{amsmath,amssymb,amsfonts}
\usepackage{algorithmic}
\usepackage{graphicx}
\usepackage{textcomp}
\usepackage{xcolor}
\def\BibTeX{{\rm B\kern-.05em{\sc i\kern-.025em b}\kern-.08em
    T\kern-.1667em\lower.7ex\hbox{E}\kern-.125emX}}
\usepackage{graphicx}
\usepackage{subfigure}
\usepackage{amsmath}
\usepackage{amsfonts}

\usepackage{amssymb}
\usepackage{enumerate}
\usepackage{booktabs}
\usepackage{framed}
\usepackage{tabularx}
\usepackage[backref]{hyperref}
\hypersetup{
colorlinks=true,
linkcolor=black,
citecolor=black,
urlcolor=black
}
\usepackage{footmisc}

\begin{document}

\title{Attention-based model for predicting question relatedness on Stack Overflow}

\author{
\IEEEauthorblockN{Jiayan Pei\IEEEauthorrefmark{1}, Yimin Wu\IEEEauthorrefmark{1, }\IEEEauthorrefmark{2}, Zishan Qin\IEEEauthorrefmark{1}, Yao Cong\IEEEauthorrefmark{1}, Jingtao Guan\IEEEauthorrefmark{2}} 
  
\IEEEauthorblockA{\IEEEauthorrefmark{1}South China University of Technology, Guangzhou, China} 
\IEEEauthorblockA{\IEEEauthorrefmark{2}Research Institute of SCUT in Yangjiang, Yangjiang, China}
\IEEEauthorblockA{seasensio@mail.scut.edu.cn, csymwu@scut.edu.cn, csqzs@mail.scut.edu.cn,\\ congyao95@hotmail.com, jingtao0337@163.com}  
}
\maketitle

\begin{abstract}
Stack Overflow is one of the most popular Programming Community-based Question Answering (PCQA) websites that has attracted more and more users in recent years. When users raise or inquire questions in Stack Overflow, providing related questions can help them solve problems. Although there are many approaches based on deep learning that can automatically predict the relatedness between questions, those approaches are limited since interaction information between two questions may be lost. In this paper, we adopt the deep learning technique, propose an Attention-based Sentence pair Interaction Model (ASIM) to predict the relatedness between questions on Stack Overflow automatically. We adopt the attention mechanism to capture the semantic interaction information between the questions. Besides, we have pre-trained and released word embeddings specific to the software engineering domain for this task, which may also help other related tasks. The experiment results demonstrate that ASIM has made significant improvement over the baseline approaches in Precision, Recall, and Micro-F1 evaluation metrics, achieving state-of-the-art performance in this task. Our model also performs well in the duplicate question detection task of AskUbuntu, which is a similar but different task, proving its generalization and robustness. 
\end{abstract}

\begin{IEEEkeywords}
Stack Overflow, Question Relatedness, Deep Learning, Attention Mechanism, Word Embeddings \end{IEEEkeywords}

\section{Introduction}
\begin{figure}[tb]
  \centering
  \includegraphics[width=0.95\linewidth]{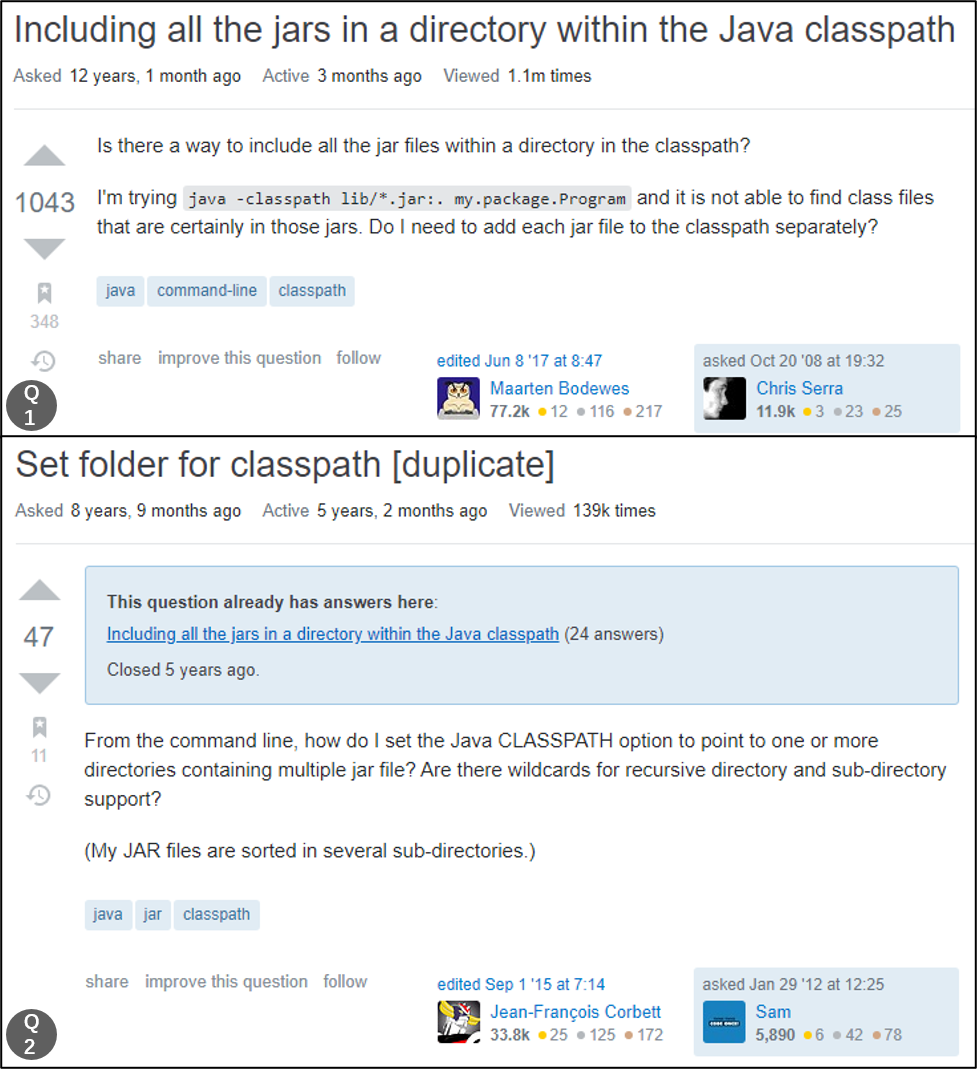}
  \caption{An example of a duplicate question pair.}
  \label{figure 1}
\end{figure}

\begin{figure}[htbp]
  \centering
  \includegraphics[width=0.95\linewidth]{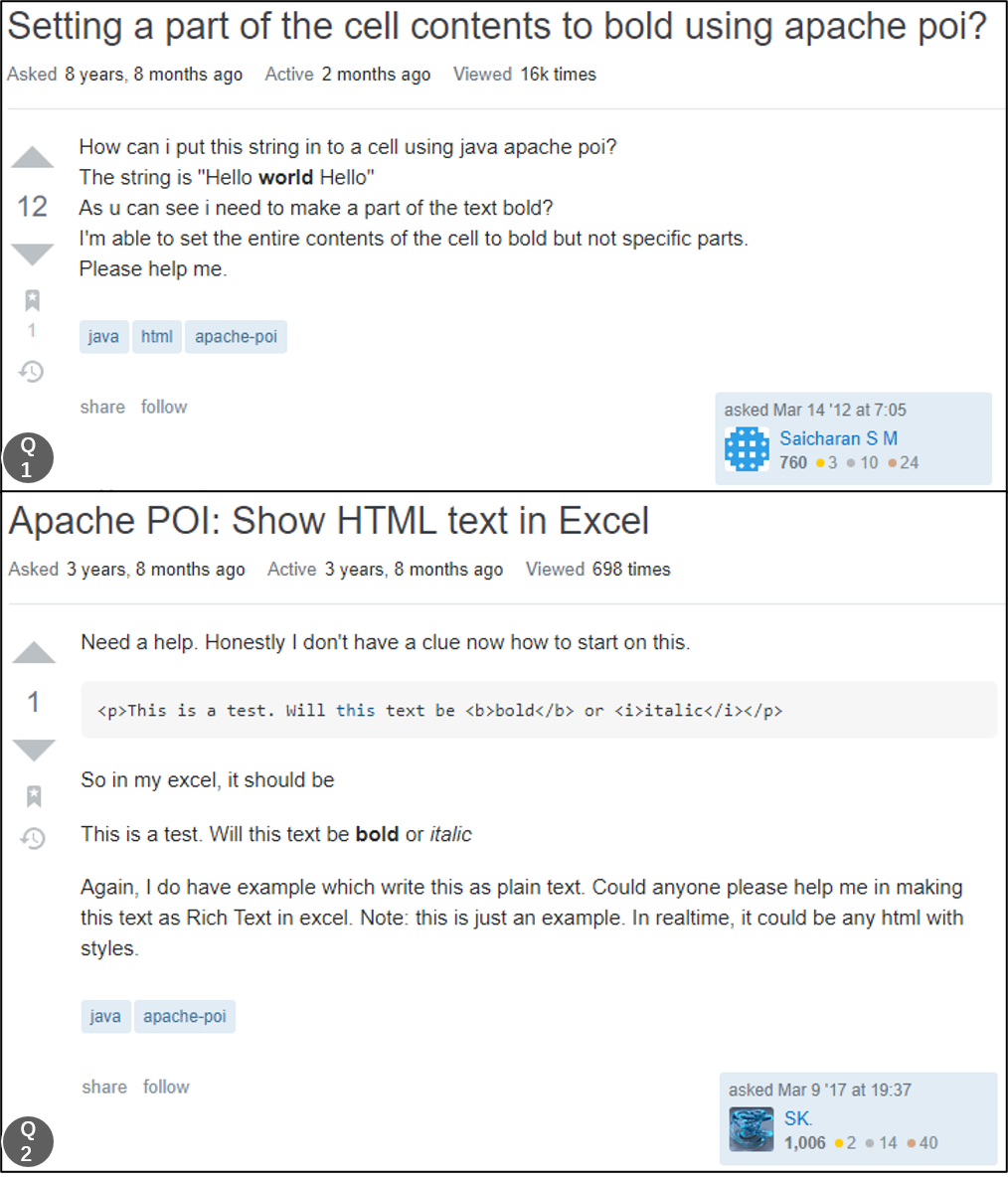}
  \caption{An example of a related question pair.}
  \label{figure 2}
\end{figure}

	The continuous progress of the information industry has caused more and more people to engage in software development. Therefore, Programming Community-based Question Answering (PCQA) websites have attracted a large number of users. Stack Overflow is one of the most popular PCQA websites, where users can ask and answer questions related to program problems and gain knowledge \cite{abdalkareem2017developers}. By October 2020, Stack Overflow had more than 20 million questions, and there could be semantic relatedness between them. For example, two or more questions may be duplicated, or information in one question may help solve other questions. Figure \ref{figure 1} presents an example of a duplicate question pair. The same solution can answer these two questions, so the recent question ('Q2') was closed and marked as '[duplicate]' and linked to the 'Q1'. Duplicate questions are not conducive to the websites' maintenance. Users who ask duplicate questions will wait a long time for the question to be answered while ready answers are already available \cite{zhang2015multi}. Figure \ref{figure 2} shows two related questions, and the answer to the question above ('Q1') can directly solve the question below ('Q2'). Therefore, providing questions related to the question raised or being inquired by users can effectively help them solve problems. However, semantic relatedness between questions often requires manual analysis by users, such as identifying and labeling duplicate questions by programmers with high reputation or indicating questions that help answer the current question in the way of URL sharing. Due to a large number of questions on the websites, and the same question can be expressed in a variety of ways, the artificial recognition method is inefficient and time-consuming, resulting in a large number of questions with semantic relatedness are not recognized. Thus, it is necessary to propose an automated approach that can help predict semantic relatedness between questions to solve this problem.

Although Stack Overflow recommends related questions when users view a question, its recommendations are based on word overlap between questions that are not reliable due to the lexical gaps \cite{xu2016predicting}. Some previous methods \cite{zhang2015multi,ahasanuzzaman2016mining,zhang2017detecting} can automatically calculate the semantic relatedness between questions, but they do not fully use the semantic information in questions \cite{wang2019detecting}. Moreover, as there are many different expressions for the same question, and there may be few words common between them, such methods' performance is poor. In recent years, deep learning has been well applied in many software engineering domains, such as user intention classification \cite{huang2018automating}, software defect prediction \cite{li2017software}, code summarization \cite{allamanis2016convolutional}, and type inference \cite{hellendoorn2018deep}. With the strong non-linear fitting ability, deep learning can also effectively extract semantic information in the question. Some previous works have adopted the deep learning technique to predict the linkable of questions \cite{xu2016predicting,liu2018linkso,amirreza2019question} or detect duplicate questions \cite{wang2019detecting,zhang2018duplicate,wang2020duplicate,bogdanova2015detecting,rodrigues2017ways} in the PCQA websites. Compared with traditional methods based on features and heuristics, their methods can achieve better results. However, most of them only adopt the sentence encoding model \cite{lan2018neural} to learn the vector representation of individual questions separately and only calculates the semantic relatedness between questions based on vector distance. Such methods can not sufficiently consider the interaction information between two questions, which is essential for semantic relatedness prediction. Recently, the attention mechanism has been widely applied in the NLP field. Using the attention mechanism to predict question relatedness, the model can extract the interaction information between questions more effectively and achieve better prediction performance through inter-sentence alignment. Therefore, in this paper, we adopt the deep learning technique to build a sentence pair interaction model based on the attention mechanism. The model aims to identify the semantic relatedness between questions on Stack Overflow.

This paper conducts research based on the work of Shirani \emph{et al.} \cite{amirreza2019question}. Their paper refers to a question and the entire set of its answers as a knowledge unit and constructs a dataset of 347,372 pairs of knowledge units (hereafter, Knowledge Unit dataset). In the Knowledge Unit dataset, the relationships between knowledge units are divided into four classes based on the degree of relatedness from high to low: \emph{duplicate}, \emph{direct}, \emph{indirect}, and \emph{isolated}. They also construct two models DOTBILSTM and SOFTSVM, based on deep learning and SVM classifier, respectively, among which the DOTBILSTM is the state-of-the-art model in this dataset. We regard the task as a multi-class classification task and proposes an Attention-based Sentence pair Interaction Model (ASIM) to solve this problem. Moreover, the most common pre-trained word embeddings are trained on a large amount of data unrelated to the software engineering domain, which may lead to ambiguous results \cite{efstathiou2018word}. Therefore, we construct a corpus of software engineering based on stack Overflow data dump and pre-train the word embeddings of domain-specific based on this corpus. Our word embeddings are openly available, which may also help design models in other software engineering tasks. In general, we answer the following five research questions in this paper:

\begin{itemize}
	\item[(1)] How much improvement can ASIM achieve over the two baseline models SOFTSVM and DOTBILSTM, to predict question relatedness?
	\item[(2)] Compared with the SOFTSVM and the DOTBILSTM, how effective is ASIM in predicting knowledge unit pairs of different relatedness classes?
	\item[(3)] How much influence does the attention mechanism contribute to the improvement of ASIM? 
	\item[(4)] What is the impact of domain-specific word embeddings on the performance improvement of ASIM?
	\item[(5)] How is the generalization ability of ASIM?
\end{itemize}

The rest of the paper is organized as follows. Section 2 briefly describes the related work of our study. Section 3  details our approaches to predict question relatedness. Section 4 describes the experimental settings and presents the experiment results. We present a case study of the attention mechanism and discuss the threats to our study's validity in Section 5, and Section 6 concludes the paper and discusses future work. 

\section{Related Work}
Semantic relatedness tasks on PCQA websites are focused areas in software engineering, including predicting semantic relatedness between questions and detecting duplicate questions.
\subsection{Predicting semantic relatedness between questions on PCQA websites}
Predicting the semantic relatedness between questions on PCQA websites is conducive to improving programmers' efficiency in finding questions and helping them solve problems using information from related questions. Xu \emph{et al.} \cite{xu2016predicting} regarded a question on Stack Overflow and the entire set of its answers as a knowledge unit. They divide the relationship between knowledge units into four classes according to the degree of relatedness: \emph{duplicate}, \emph{direct}, \emph{indirect}, and \emph{isolated}. They adopt a deep learning approach, which encodes the vector representation of two knowledge units respectively through a convolutional neural network (CNN) with shared parameters. They then calculate the cosine similarity of the two feature vectors extracted from the knowledge unit pairs to obtain the semantic similarity. The results are predicted to one of four classes mentioned above according to the semantic similarity. Compared with the SVM-based model, this neural network model can achieve better results. To the best of our knowledge, Xu \emph{et al.} are the first to apply the deep learning technique to predict semantic relatedness between questions on PCQA websites. Based on the work of Xu \emph{et al.}, Shirani \emph{et al.} \cite{amirreza2019question} built a dataset containing more than 300,000 pairs of knowledge units. They constructed two baseline models based on the deep learning technique and machine learning technique, respectively. The deep learning model DOTBILSTM encodes a pair of knowledge units based on bidirectional LSTM, calculates the inner dot product of three parts (title, body, and answer) between a pair of knowledge units, maps it to a low-dimensional vector space. The vector is then inputted into a fully-connected layer using dense connection and  a softmax output layer for classification. SOFTSVM is a Support Vector Machine (SVM) model based on the soft-cosine similarity features of knowledge unit pairs. The experimental results show that DOTBILSTM can achieve better performance on this dataset. Liu \emph{et al.} \cite{liu2018linkso} also studied the semantic relatedness between questions. They construct the LinkSO dataset based on the links between the questions from the Stack Overflow data dump. This dataset contains three different programming language questions and a total of 26,593 linked question pairs. The manual sampling analysis of the LinkSO dataset shows that most of the linked questions are relevant, so the dataset may help with studying the semantic relatedness between questions on Stack Overflow. They compare the performance of six existing models on this dataset. Three approaches are non-learning approaches (TF-IDF, BM25, TransLM), and the other three are learning-based approaches (DSSM, DRMM, aNMM). Experiment results show that the learning-based approaches can achieve better performance, among which the aNMM is the best-performing approach on this dataset.

In the previous work, most deep learning approaches were based on the sentence  encoding model\cite{lan2018neural}, which learns the vector representation of a single question and only calculates the semantic relevance between questions based on the vector distance in the high layer. These methods can not sufficiently consider the interaction information between two questions, which is essential to predicting the semantic relatedness. Therefore, based on the deep learning method, ASIM adopts the attention mechanism to learn the semantic interaction between two questions, hoping to achieve better performance in predicting question relatedness on PCQA websites.
\subsection{Duplicate question detection in PCQA websites}
There are numerous unrecognized duplicate questions in the PCQA websites, which are not conducive to websites' maintenance. Moreover, users who ask duplicate questions wait a long time for the question to be answered, while ready answers are already available \cite{zhang2015multi}. Although some PCQA websites such as Stack Overflow recommend users search for related questions before posing new ones \cite{ahasanuzzaman2016mining}, as the same questions can be expressed in many different ways, the website will inevitably face duplicate questions. Taking Stack Overflow as an example, users of high reputation mark duplicate questions through manual analysis \cite{ahasanuzzaman2016mining}, which is ineffective and time-consuming. Therefore, automated detection approaches are required. Zhang \emph{et al.} \cite{zhang2015multi} proposed a tool DUPPREDICTOR for automatically detecting duplicate questions on Stack Overflow. The tool uses Latent Dirichlet Allocation (LDA) to transform the natural language in a question into a topic distribution. Then it calculates the semantic similarity of the question pair by considering the title, description, topic, and tag similarity. This tool is the first work that addresses the problem of duplicate questions on Stack Overflow. Ahasanuzzaman \emph{et al.} \cite{ahasanuzzaman2016mining} proposed a duplicate question detection model Dupe for Stack Overflow. This model uses logistic regression to detect the duplicate question based on the five features of the question pair, such as cosine similarity value, term overlap, entity overlap, entity type overlap, and WordNet similarity. In the prediction stage, the BM25 algorithm and other methods are applied to filter irrelevant questions. Although the above methods can automatically detect duplicate questions, they require the manual design of features, which is time-consuming, and do not fully use the semantic information in question \cite{wang2019detecting}. Moreover, these features are usually related to specific tasks, so such methods' generalization performance is lacking.

Wang \emph{et al.} \cite{wang2019detecting} first applied the deep learning technique to detect duplicate questions on Stack Overflow. They explore three different deep learning approaches (e.g., DQ-CNN, DQ-RNN, and DQ-LSTM) based on Convolutional Neural Network (CNN), Recurrent Neural Networks (RNN), and Long Short-Term Memory (LSTM), respectively. Compared with the baseline detection approaches and machine learning approaches, their models can achieve better performance, indicating the superiority of deep learning in this field. Besides, the DQ-LSTM model based on LSTM performs better than other deep learning models. AskUbuntu is another popular CQA website focused on software programming. For the duplicate question detection on the AskUbuntu, Bogdanova \emph{et al.} \cite{bogdanova2015detecting} applied the convolution neural network with shared parameters to encode the distributed vector representation of individual questions and then calculate the similarity of vectors through cosine similarity to predict whether a pair of questions are duplicate. Based on Bogdanova \emph{et al.}'s work, Rodrigues \emph{et al.} \cite{rodrigues2017ways} released the AskUbuntu dataset's clean version. They removed the explicit clues from the question to avoid a biased result \cite{silva201820}. In their paper, the DCNN model, which combines the CNN model \cite{bogdanova2015detecting} and the MayoNLP model \cite{afzal2016mayonlp}, can achieve state-of-the-art performance on this dataset. Compared with the traditional automated detection methods, the above deep learning approaches do not require the manual design of features and have better performance and generalization ability. However, such methods also fail to consider the interaction information between two questions well. We also expand our approach to the duplicate question detection task in software engineering, using the clean version of the AskUbuntu dataset prepared by Rodrigues \emph{et al.} \cite{rodrigues2017ways} to evaluate ASIM's applicability and robustness in a similar task.

\section{The Approach}
According to Shirani \emph{et al.} \cite{amirreza2019question}, a question on Stack Overflow and the entire set of its answers is a knowledge unit (KU). Based on the degree of relatedness between two knowledge units from high to low, the relatedness types between them can be defined as the following four classes:
\begin{itemize}
\item \emph{duplicate}: The questions in the two knowledge units are duplicate questions.
\item \emph{direct}: Information in one knowledge unit can directly solve the question in another knowledge unit.
\item \emph{indirect}: The information in one knowledge unit is helpful to the solution of the question in another knowledge unit, but the information alone cannot directly solve the question.
\item \emph{isolated}: There is no semantic relatedness between the two knowledge units.
\end{itemize}

We treat the task as a multi-class classification problem. The model's input is a pair of knowledge units, and the relatedness is predicted to be one of the four classes mentioned above.

\begin{figure}[tb]

\centering
\includegraphics[scale=0.22]{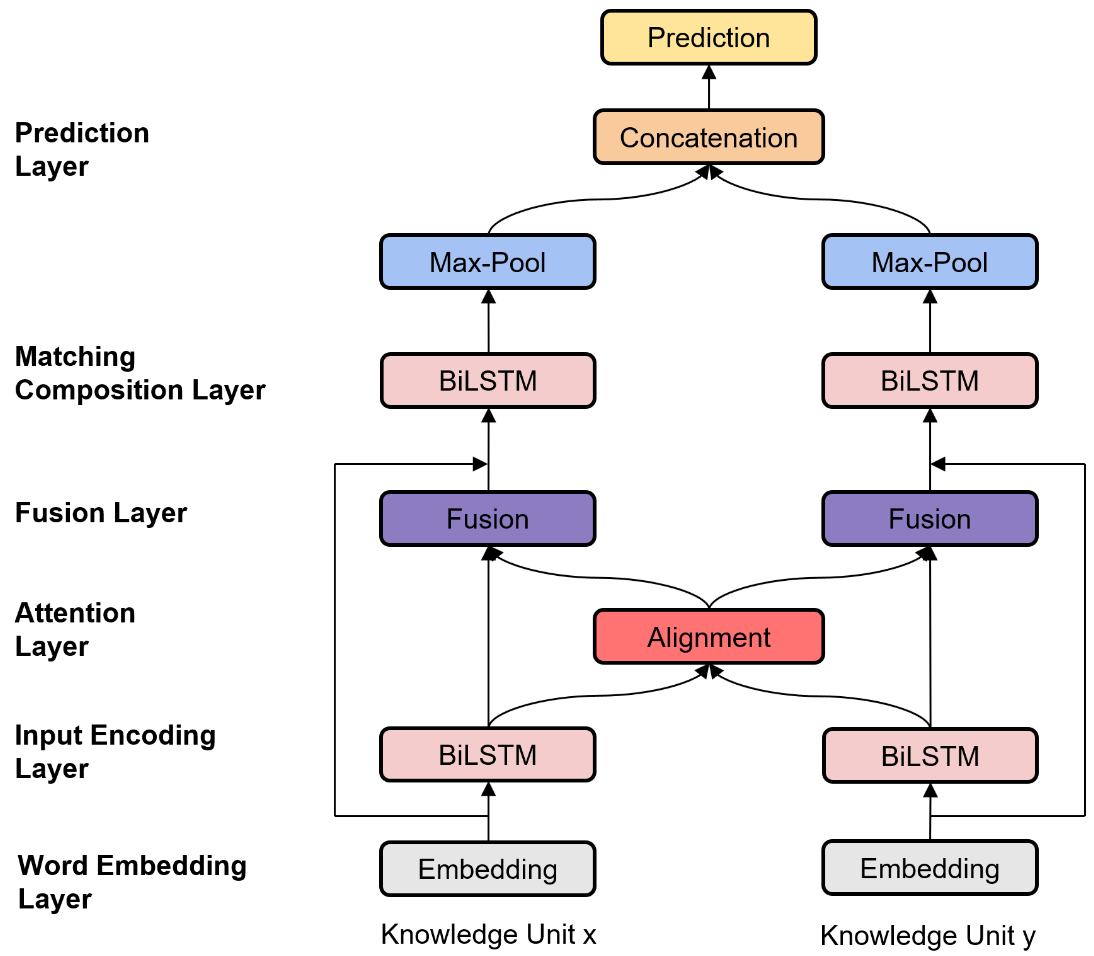}
\caption{The framework of ASIM.}
\label{fig:model}
\end{figure}

Figure \ref{fig:model} gives an illustration of the ASIM framework, which is mainly composed of the following seven modules: (1) Word Embedding Layer, (2) Shortcut Connections, (3) Input Encoding Layer, (4) Attention Layer, (5) Fusion Layer, (6) Matching Composition Layer, and (7) Prediction Layer.
\subsection{Input of the model}
We concatenate the title and body of the question and the set of the answer to form the text sentence of a knowledge unit. Suppose the text sequence of a knowledge unit is $\{w_1, w_2, ..., w_n\}$, where n is the sequence length. The ASIM's inputs are $KUx$ and $KUy$ and the label of relatedness, among which $KUx$ and $KUy$ are text sequences of a pair of knowledge units. Since the two text sequences are treated symmetrically before the Prediction Layer, and all parameters except the Prediction Layer are shared between the two sequences. For brevity, we will only introduce the processing method of $KUx$ in the model.
\subsection{Word Embedding Layer}
In NLP tasks, words are usually transferred into the corresponding vector representations by word embeddings \cite{mikolov2013distributed}. Word embeddings are trained on a large number of unlabeled texts through unsupervised learning, which can capture the rich semantic and syntactic information of words. Shirani \emph{et al.} \cite{amirreza2019question} carried out experiments applying the pre-trained GloVe word embeddings released by Stanford \cite{pennington2014glove}. These word embeddings pre-trained in the Common Crawl corpus, which contains a large amount of data irrelevant to software engineering, may lead to ambiguous results \cite{efstathiou2018word}. Therefore, we hope that there will be word embeddings specific to the software engineering domain for this task, but also the research of other tasks in software engineering. We build a 15GB corpus based on Stack Overflow data dump, where the data dump spans a temporal interval from January 2014 to June 2020. Specifically, we used the posts part to extract the content in the body tag, and clean the content by removing the HTML tag, code snippets, URL link, punctuation, stop words, and changing all words to lowercase and stemming. Finally, the GloVe method \cite{pennington2014glove} was adopted for pre-training to obtain word embeddings\footnote{Our pre-trained word embeddings are openly available in \url{https://zenodo.org/record/4641569}} with dimensions of 300.

In Word Embedding Layer, each word in the knowledge unit, namely $w_{i}$, is transformed into a vector representing $x_{i} \in \mathbb{R}^{d}$ through our pre-trained word embeddings. Therefore, knowledge units containing $n$ words can be converted into corresponding matrix representation:
\begin{equation}
KUx = {x_1} \oplus {x_2} \oplus  \cdots  \oplus {x_n}
\end{equation}
Where $\oplus$ is the concatenation operator, and the result is the input of the next layer of the model.
\subsection{Shortcut Connections}
Between the Input Encoding Layer and the Attention Layer, and between the Matching Composition Layer and the Prediction Layer, we adopt the approach similar to residual connection \cite{he2016deep}, which can effectively mitigate the gradient vanishing and gradient exploding and retain the original features to some extent. Different from the traditional residual connection, we refer to the idea of Huang \emph{et al.} \cite{huang2017densely} and Kim \emph{et al.} \cite{kim2019semantic}, change the summation operation to concatenation operation so that the characteristics of the previous layer cannot be modified and better retained:
\begin{equation}
O^l = {H_l}({I^l}),\quad I^l = [{O^{(l - 2)};O^{(l - 1)}}]
\end{equation}
Where $O^l$ represents the output of the $l^{th}$ layer $H_l$, and $I^l$ represents the input of the layer $H_l$, and $[;]$ denotes the concatenation operation. Unlike the previous layers, for the Matching Composition Layer's input, we refer to the idea of Yang \emph{et al.} \cite{yang2019simple}, adopt another version of the residual connection. As shown in Figure \ref{fig:model}, between the Matching Composition Layer and the Fusion Layer, the Attention Layer's output is not connected. However, the word vector representation corresponding to the original knowledge unit is connected to better retain the original meaning of the word in the high layer:
\begin{equation}
O^l = {H_l}({I^l}),\quad I^l = [ {O^{(l - 1)};KNx}]
\end{equation}
\subsection{Input Encoding Layer}
In this layer, we adopt BiLSTM to fuse contextual information into each word's original representation. BiLSTM is composed of a forward and a backward LSTM \cite{hochreiter1997long}. Through its three gate structures, LSTM can solve the long-term dependence problem very well. With BiLSTM, bi-directional semantic dependencies within the knowledge unit can be well captured. Precisely, one LSTM captures information from the first time step to the last time step, and the other LSTM captures information in reverse. Then we connect the outputs of the two LSTM to obtain the augmented representation of the word, which incorporates the contextual information. The following formula can represent this step:
\begin{gather}
{\vec h_i} = LSTM({\vec h_{i - 1}},{x_i}), \forall i \in [1,...,n]\\
{\mathord{\buildrel{\lower3pt\hbox{$\scriptscriptstyle\leftarrow$}} 
\over h} _i} = LSTM({\mathord{\buildrel{\lower3pt\hbox{$\scriptscriptstyle\leftarrow$}} 
\over h} _{i - 1}},{x_i}), \forall i \in [1,...,n]\\
{X_i} = [{\vec h_i};{\mathord{\buildrel{\lower3pt\hbox{$\scriptscriptstyle\leftarrow$}} 
\over h} _i}]
\end{gather}
Where $\vec h_{i - 1}$ is the hidden layer state of the forward LSTM in the time step $i-1$, ${\mathord{\buildrel{\lower3pt\hbox{$\scriptscriptstyle\leftarrow$}} 
\over h} _{i - 1}}$ is the hidden layer state of the backward LSTM, $x_i$ is the input of LSTM in the time step $i$, and $X_i$ is the contextual representation of word $w_i$.
\subsection{Attention Layer}
The most previous deep learning models used in semantic relatedness tasks of PCQA websites are sentence encoding models. These methods do not consider the interaction information between the two questions. In the field of NLP, attention mechanisms were first used in the neural machine translation model \cite{bahdanau2014neural}. By applying the attention mechanism to the question relatedness prediction task, the model can effectively extract the interaction information between questions through inter-sentence alignment and achieve better performance. 

In this paper, the attention mechanism is applied to capture the knowledge units' semantic interaction. For $X_i$ and $Y_j$ ($i \in [1, n], j \in [1, m]$), that is, the output of two words in the knowledge unit $KUx$ and $KUy$ from the Input Encoding Layer, first calculate the Inter-Attention matrix $E \in {\mathbb{R}^{n \times m}}$:
\begin{equation}
{E_{i,j}} = X_i^T \cdot {Y_j}\label{equation 7}
\end{equation}
Where $\cdot$ denotes the inner production operation, and the $i-th$ row $j-th$ column in the Inter-Attention matrix represents relevance between the $i-th$ word in $KUx$ and the $j-th$ word in $KUy$. Based on the Inter-Attention matrix, we can calculate the Inter-Attention vector ${e_{{X_i}}}$ and ${e_{{Y_j}}}$, which represent the semantic correlation between a word and another knowledge unit:
\begin{gather}
{e_{{X_i}}} = softmax(\frac{{{E_{i,:}}}}{{\sqrt k }})\\
{e_{{Y_j}}} = softmax(\frac{{{E_{:,j}}}}{{\sqrt k }})
\end{gather}
It is worth noting that before softmax, we referred to the idea of the paper \cite{vaswani2017attention} and divided the Inter-Attention matrix by $\sqrt k$, where $k$ is the dimension of vector $X_i$ and $Y_j$. This method can make the gradient more stable. Finally, we can get the Inter-Attention representation $\hat X_i$ and $\hat Y_j$:
\begin{gather}
{\hat X_i} = Y \cdot {e_{{X_i}}}\\
{\hat Y_j} = X \cdot {e_{{Y_j}}}
\end{gather}
Where $\hat X_i$ is the vector representation of the $i-th$ word in $KUx$ that fuses the interaction information with $KUy$, $\hat Y_i$ is the vector representation of the $j-th$ word in $KUy$ that fuses the interaction information with $KUx$.
\subsection{Fusion Layer}
In the Fusion Layer, the contextual representation and the Inter-Attention representation of word are integrated to fuse aligned features. We refer to the fusion method in the paper \cite{yang2019simple} and use the following three ways to fuse the features:
\begin{gather}
\tilde X_i^1 = {F_1}([{X_i};{\hat X_i}])\\
\tilde X_i^2 = {F_2}([{X_i};{X_i} - {\hat X_i}])\\
\tilde X_i^3 = {F_3}([{X_i};{X_i} \circ {\hat X_i}])
\end{gather}
Where $-$ denotes calculate the difference between vectors, $\circ$ represents element-wise multiplication, and $[;]$ refers to the concatenation operation. The subtraction operator highlights the difference between two vectors, while the multiplication highlights the similarity between the vectors \cite{yang2019simple}. $F1$, $F2$, $F3$ are three single-layer feed-forward networks with independent parameters. Then we concatenate the results obtained by the above three fusion methods and input them into another single-layer feed-forward network $F$ to compute the output of the Fusion Layer:
\begin{equation}
{\tilde X_i} = F([\tilde X_i^1;\tilde X_i^2;\tilde X_i^3])
\end{equation}
Similarly, for the knowledge unit $KUy$, the same method is applied to obtain the Fusion Layer's output $\tilde Y$.
\subsection{Matching Composition Layer}
In the Matching Composition Layer, we use another BiLSTM to extract key information from the output of the Fusion Layer to obtain the final vectors $V_i^X$ and $V_j^Y$ of the two knowledge units:
\begin{gather}
V_i^X = BiLSTM(\tilde X,i),\forall i \in [1,...,n]\\
V_j^Y = BiLSTM(\tilde Y,j),\forall j \in [1,...,m]
\end{gather}
Where $n$ and $m$ are the length of knowledge units $KUx$ and $KUy$, respectively.
\subsection{Prediction Layer}
In the Prediction Layer, for the vectors obtained from the Matching Composition Layer, we use max-pooling to convert them into fixed-size vectors:
\begin{gather}
V_{\max }^X = \mathop {\max }\limits_{i = 1}^n v_i^X\\
V_{\max }^Y = \mathop {\max }\limits_{j = 1}^m v_j^Y
\end{gather}

Then we use a method similar to the paper \cite{yang2019simple,mou2015natural} to concatenate the original vector, the difference between the two vectors, and the result of element-wise multiplication, and input it into a multi-layer feed-forward network $L$ to get the feature vector:
\begin{equation}
V = L([V_{\max }^X;V_{\max }^Y;V_{\max }^X - V_{\max }^Y;V_{\max }^X \circ V_{\max }^Y])
\end{equation}

Finally, the softmax function is applied to obtain the probability distribution of each class.

\section{Experiment}
\subsection{Dataset and Setup}
We carry out experiments on the Knowledge Unit dataset\footnote{\url{https://anonymousaaai2019.github.io}} built by Shirani \emph{et al.} \cite{amirreza2019question}. This dataset is constructed based on Stack Overflow data dump and contains 347,372 pairs of knowledge units. Moreover, all the knowledge units in this dataset are Java-related because Java is one of the top-3 most popular tags in Stack Overflow. In this dataset, knowledge unit pairs have four relatedness classes, and the number of each relatedness class accounts for 1/4. 60\% of the dataset's knowledge unit pairs are used as a training set, 10\% as a validation set, and 30\% as a test set. We randomly pick knowledge unit pair samples of each class in the dataset and list them in Table \ref{table 1}.

\begin{table*}
\caption{Four examples from the Knowledge Unit dataset.}
\footnotesize
\centering
\begin{tabular}{lll}
\toprule
\textbf{Lable} & \textbf{Knowledge Unit x} & \textbf{Knowledge Unit y} \\ \midrule
\emph{\textbf{duplicate}} & \begin{tabular}[t]{@{}p{7.5cm}@{}}\textbf{Question Id:} 36734301\\ \textbf{Title:} How to declare a call a 2d array in java\\ \textbf{Body:} I am trying to read an image's pixels and fill them in a 2d array however I do not know how to declare a global array any help please\\ \textbf{Answers:} null\end{tabular} & \begin{tabular}[t]{@{}p{7cm}@{}}\textbf{Question Id:} 19894714\\ \textbf{Title:} How can I create 2D arrays in java\\ \textbf{Body:} How would I go about designing something like this using 2D arrays in java Everything works but name i j = 200 when i put this it only prints this and nothing else\\ \textbf{Answers:} You would replace name with what you would like to name the array and you would replace x and y with the x and y \emph{(80 more words omitted)}\end{tabular} \\ \midrule
\emph{\textbf{direct}} & \begin{tabular}[t]{@{}p{7.5cm}@{}}\textbf{Question Id:} 8147454\\ \textbf{Title:} how to instantiate a class while jboss startup\\ \textbf{Body:} I would like to instantiate my own java class One time only when the time of startup of JBOSS 5 and i will use that object until i shut down the jboss How can it be possible to instantiate.\\ \textbf{Answers:} You can implement your class with the ServletContextListener interface which make your class able to receive notifications from the application server \emph{(42 more words omitted)}\end{tabular} & \begin{tabular}[t]{@{}p{7cm}@{}}\textbf{Question Id:} 36495078\\ \textbf{Title:} Java execute method when .war file is deployed\\ \textbf{Body:} I want to execute some methods as soon as the .war file is deployed by Tomcat or JBoss how can I do it I tried ServletContextListener but it's not working Thanks.\\ \textbf{Answers:} 'Have you tried adding your methods in a Servlet and then run it on startup In your Web.xml ', 'OK I resolved this this works with JBoss And this works with Tomcat '\end{tabular} \\ \midrule
\emph{\textbf{indirect}} & \begin{tabular}[t]{@{}p{7.5cm}@{}}\textbf{Question Id:} 4858022\\ \textbf{Title:} How to configure a log4j file appender which rolls the log file every 15 minutes\\ \textbf{Body:} I understand that i can use a DailyRollingFileAppender to roll the log file every month day half-day hour or minute But how can i configure log4j to roll the log file every 15 minutes If this is not possible by configuration please suggest direct me on how to extend log4j's file appender to achieve this Thanks and Regards.\\ \textbf{Answers:} The Javadoc for DailyRollingFileAppender in Log4J indicates that the time-based rolling only occurs on unit-based rollovers day week month \emph{(319 more words omitted)}\end{tabular} & \begin{tabular}[t]{@{}p{7cm}@{}}\textbf{Question Id:} 5030536\\ \textbf{Title:} How to log impressions and data in Java for a javascript widget\\ \textbf{Body:} I have a javascript widget the loads JSON data from a Java webapp I want to record impressions and the ids of the data I return 5 or 10 longs \emph{(97 more words omitted)}\\ \textbf{Answers:} Writing to log and processing off-line should be ok You can program your logging system to create hourly log files then process files that are not written-to any more \emph{(62 more words omitted)}\end{tabular} \\ \midrule
\emph{\textbf{isolated}} & \begin{tabular}[t]{@{}p{7.5cm}@{}}\textbf{Question Id:} 25579459\\ \textbf{Title:} Got apache tomcat error that access denied on this file localhost\_access\_log.2014-08-30.txt\\ \textbf{Body:} I installed Apache tomcate on Windows 7 OS I just installed apache and make it available in eclipse When i run any simple application on server it will say 404 page not found and in console it will print the error message like this I cannot understand what the things is happen is this Please help me.\\ \textbf{Answers:} It seems that your current user is not having rights on the tomcat folder I also faced the same problem and solved it by giving rights to the logged in user on the tomcat folder \emph{(66 more words omitted)}\end{tabular} & \begin{tabular}[t]{@{}p{7cm}@{}}\textbf{Question Id:} 3690114\\ \textbf{Title:} Set Depth java.util.TreeMap\\ \textbf{Body:} How can we set depth of a TreeMap object Suppose we are trying to build an auto suggest feature on top of underlying data structure of a TreeMap how would depth of a tree as we know affect the performance\\ \textbf{Answers:} Your question is vague but if I understand correctly you're misunderstanding concepts TreeMap is an implementation of the Map interface which uses red-black tree for sorting its contents into natural ascending order while what you're asking is something completely unrelated; \emph{(99 more words omitted)}\end{tabular} \\ \bottomrule
\end{tabular}
\label{table 1}
\end{table*}

\subsection{Evaluation Metric}
We use the same evaluation metrics as Shirani \emph{et al.} \cite{amirreza2019question} to evaluate ASIM's performance, including Precision, Recall, and Micro-F1.

\textbf{Precision} represents the proportion of samples predicted to be positive that are truly positive samples. Precision for all classes is the mean
of the precision for each class.
\begin{equation}
Precision = \frac{{Tru{e^{}}Positive}}{{Tru{e^{}}Positive + Fals{e^{}}Positive}}
\end{equation}

\textbf{Recall} represents the proportion of the positive samples that are correctly predicted to be positive samples. Recall for all classes is the mean of the recall for each class.
\begin{equation}
Recall = \frac{{Tru{e^{}}Positive}}{{Tru{e^{}}Positive + Fals{e^{}}Negative}}
\end{equation}

\textbf{F1-score} is the weighted average of Precision and Recall. This metric takes into account both the Precision and Recall of the model:
\begin{equation}
F1 = 2 \cdot \frac{{precision \cdot recall}}{{precision + recall}}
\end{equation}

In the multi-class classification task, precision and recall are calculated by considering all classes together and measures the F1-score of all classes' aggregated contributions to get the Micro-F1.
\subsection{Implementation Details}
ASIM is implemented based on the PyTorch \cite{paszke2017automatic} framework, references the work of paper \cite{yang2019simple} and experimented on an Nvidia 1080Ti GPU. The input of ASIM is a knowledge unit pair and its label: $<KUx, KUy, label>$. We apply some data pre-processing steps on the input text, including normalizing URLs and numbers, removing punctuation marks and stop-words, splitting camel case words, stemming, and changing all words to lowercase. The knowledge unit's maximum length is set to 250 to keep it consistent with the paper \cite{amirreza2019question}. We initialize word embeddings with our pre-trained word vector  of size 300 and fixed it during training. Compared to GloVe word embeddings, the word embeddings specific to the software engineering domain achieve better performance in this task. Adam optimizer \cite{kingma2014adam} with an initial learning rate of 0.0012 was applied. For the Input Encoding Layer and Matching Aggregation Layer, we use BiLSTM with 200 units as the encoder. The dropout strategy \cite{srivastava2014dropout} is adopted at each BiLSTM and fully-connected layer, and the dropout rate is set to 0.2. We use a batch size of 128. The model is trained for 30 epochs to minimize the cross-entropy loss, and the validation set of the Knowledge Unit dataset was used for the model's hyper-parameters tuning. We release the source code of our model\footnote{Our model is openly available in \url{https://github.com/Anonymousmsr/ASIM}} for more details and hope to facilitate future researches.
\subsection{Research Questions}
We are interested in answering the following research questions:

\textbf{RQ1:} \emph{\textbf{How much improvement can ASIM achieve over the two baseline models SOFTSVM and DOTBILSTM, to predict question relatedness?}}

\textbf{Motivation.} ASIM adopts the deep learning technique and can better learn the interaction information between knowledge unit pairs based on the attention mechanism, which is quite different from the SOFTSVM and the DOTBILSTM.  Moreover, the DOTBILSTM is the state-of-the-art model on the Knowledge Unit dataset. The answer to this research question will shed light on whether and to what extent ASIM can improve the results on predicting question relatedness on Stack Overflow.

\textbf{Approach.} We compare the performance of ASIM on the test set with the SOFTSVM and the DOTBILSTM. We evaluate different models on Precision, Recall, and Micro-F1 metrics. The results of the SOFTSVM and the DOTBISLTM are from the paper \cite{amirreza2019question}, which keeps the results to two decimal places. To present the results more accurately, we keep the results of ASIM to four decimal places.

\textbf{Result.} Table \ref{table 2} presents the experiment result. We can see that compared with DOTBILSTM and SOFTSVM, ASIM has improved by 7\% and 23\%, respectively, in terms of the Micro-F1. There is a similar improvement to the Precision and Recall.
\begin{table}
\caption{Micro-F1, Precision, and Recall of ASIM and the baseline models.}
\centering
\begin{tabular}{llll}
\toprule
\textbf{Model/Metrics} & \textbf{Micro-F1} & \textbf{Precision} & \textbf{Recall} \\
\midrule
SOFTSVM       & 0.59    & 0.58      & 0.59   \\
DOTBILSTM     & 0.75    & 0.75      & 0.75   \\
\midrule
ASIM     & \textbf{0.8228}    & \textbf{0.8210}      & \textbf{0.8228}   \\
\bottomrule
\end{tabular}
\label{table 2}
\end{table}

\begin{framed}
\textbf{RQ1:} The experiment result shows that ASIM, based on attention mechanism, outperforms the baseline approaches SOFTSVM and DOTBILSTM in all the evaluation metrics and achieves state-of-the-art performance in the Knowledge Unit dataset.
\end{framed}
\textbf{RQ2:} \emph{\textbf{Compared with the SOFTSVM and the DOTBILSTM, how effective is ASIM in predicting knowledge unit pairs of different relatedness classes?}}

\textbf{Motivation.} The semantic relatedness between the knowledge unit pairs of different classes is inconsistent, and the prediction difficulty of the model is also different. We hope to compare ASIM's prediction performances with that of two baseline models on each class of knowledge unit to study the improvement and advantages of ASIM better.

\textbf{Approach.} We compare ASIM's prediction results on each knowledge unit class with the SOFTSVM and DOTBILSTM on the F1-score metric. The results of the two baseline models are from the paper \cite{amirreza2019question}.

\textbf{Result.}  It can be seen from Table \ref{table 3} that ASIM outperforms DOTBiLSTM and SOFTSVM in all four classes. For the \emph{duplicate} class, because there is a high degree of semantic relatedness between the two knowledge units, the DOTBILSTM's performance is well in this class. Hence, ASIM has a slight improvement (1\%) compared with DOTBILSTM. For the remaining three classes, ASIM's improvement is more significant (5\%-13\%) compared with DOTBILSTM. However, there is a certain degree of semantic relatedness between knowledge units of \emph{direct} and \emph{indirect} classes, but the distinction between these two classes is no obvious. Therefore, ASIM performs worse in these two classes than \emph{duplicate} and \emph{isolated} classes, and these two classes are also bottlenecks to limit the overall performance of ASIM. For the \emph{isolated} class, ASIM can predict well and achieve competitive performance because there is no semantic relatedness between the knowledge units, which can be well recognized.
\begin{table}
\caption{Comparing the result (F1-score) of ASIM and the baseline models in each relatedness class.}
\centering
\resizebox{250pt}{9.5mm}{
\begin{tabular}{llllll}
\toprule
\textbf{Model/Classes} & \textbf{duplicate} & \textbf{direct} & \textbf{indirect} & \textbf{isolated} & \textbf{Micro-F1} \\
\midrule
SOFTSVM       & 0.53      & 0.57   & 0.44     & 0.79     & 0.59          \\
DOTBILSTM     & 0.92      & 0.55   & 0.67     & 0.87     & 0.75          \\
\midrule
ASIM     & \textbf{0.9299} & \textbf{0.6845} & \textbf{0.7284} & \textbf{0.9437} &\textbf{0.8228}\\    
\bottomrule
\end{tabular}}
\label{table 3}
\end{table}

\begin{framed}
\textbf{RQ2:} ASIM outperforms the two baseline models on all four classes. Besides, for \emph{direct}, \emph{indirect}, and \emph{isolated} classes, ASIM's improvement is more significant.
\end{framed}
\textbf{RQ3:} \emph{\textbf{How much influence does the attention mechanism contribute to the improvement of ASIM?}}

\textbf{Motivation.} The attention mechanism is a crucial part of ASIM, and it helps the model capture the semantic interaction between knowledge units, which is quite different from the previous models. The answer to this research question helps us understand the importance of the attention mechanism to the ASIM.

\textbf{Approach.} We remove the Attention Layer (Attn) and compare the revised model's performance with the original model on the F1-score. Specifically, we remove the Attention Layer and the Fusion Layer simultaneously and remove the related Shortcut Connections. Although this setup may show the layer importance, it is unclear whether the Fusion Layer or the Attention Layer or the Shortcut Connection is the most important. Thus, we also perform the following ablation studies, consisting in (1) only removing Fusion Layer (FL) and replacing by a concatenation operation; (2) Only removing the Shortcut Connections (SC) related to the Attention Layer and (3) removing both the Fusion Layer and the Shortcut Connections. Then, we measure how much these isolated modifications impact model's performance.

\textbf{Result.} The experimental results are shown in Table \ref{table 4}. After removing the Fusion Layer or Shortcut Connection, respectively, or removing both of them, the revised models' performance decreases slightly (0.9\%, 0.6\%, and 1.0\% in terms of Micro-F1, respectively). However, after removing the Attention Layer, the model's performance decreases more obviously (3.8\% in Micro-F1), especially in predicting the \emph{direct} and the \emph{indirect} classes. This experiment proves the importance of the attention mechanism, which plays an essential role in predicting questions relatedness. The attention mechanism's introduction has a more prominent effect on the \emph{direct} class and the \emph{indirect} class. Furthermore, even if the attention mechanism is removed, our model's performance still outperforms the two baseline models.
\begin{table}
\caption{Comparing the result (f1-score) of ASIM and the revised models in each relatedness class.}
\centering
\resizebox{250pt}{13mm}{
\begin{tabular}{p{65pt}ccccc}
\toprule
\textbf{Model/Classes}                    & \textbf{duplicate} & \textbf{direct} & \textbf{indirect} & \textbf{isolated} & \textbf{Micro-F1} \\
\midrule
\textbf{ASIM}                        & \textbf{0.9299}      & \textbf{0.6845}   & \textbf{0.7284}     & \textbf{0.9437}     & \textbf{0.8228}          \\
ASIM ( - FL)             & 0.9249      & 0.6552   & 0.7245     & 0.9404     & 0.8137          \\
ASIM ( - SC)             & 0.9222      & 0.6838   & 0.7175     & 0.9384     & 0.8165          \\
ASIM ( - FL - SC)             & 0.9177      & 0.6745   & 0.7097     & 0.9384     & 0.8124          \\
ASIM ( - Attn - FL - SC)             & 0.9153      & 0.6251   & 0.6697     & 0.9217     & 0.7846          \\
\bottomrule
\end{tabular}}
\label{table 4}
\end{table}

\begin{framed}
\textbf{RQ3:} The attention mechanism improves the performance of ASIM to some extent. Moreover, the attention mechanism plays an essential role in the prediction of \emph{direct} and \emph{indirect} classes.
\end{framed}

\begin{table}[b]
\caption{Performance (F1-score) of ASIM with different Word Embeddings.}
\centering
\resizebox{250pt}{9mm}{
\begin{tabular}{p{70pt}ccccc}
\toprule
\textbf{Model/Classes}                 & \textbf{duplicate} & \textbf{direct} & \textbf{indirect} & \textbf{isolated} & \textbf{Micro-F1} \\
\midrule
GloVe word embeddings          & 0.9242      & 0.6667   & 0.7262     & 0.9365     & 0.8154          \\
Doman-specific word embeddings & \textbf{0.9299}      & \textbf{0.6845}   & \textbf{0.7284}     & \textbf{0.9437}     & \textbf{0.8228}      \\
\bottomrule
\end{tabular}}
\label{table 5}
\end{table}

\begin{table*}
\caption{Two example question pairs and their labels from the AskUbuntu dataset.}
\footnotesize
\centering
\begin{tabular}{lll}
\toprule
\textbf{Lable} & \textbf{Question 1} & \textbf{Question 2} \\ \midrule
\emph{\textbf{duplicate}} & \begin{tabular}[t]{@{}p{7.2cm}@{}}\textbf{Title:} Where can I find the source code of Ubuntu?\\ \textbf{Body:} I would like to know where to find the source code of Ubuntu 12.04. I'd like to see how far it is "open source".\end{tabular} & \begin{tabular}[t]{@{}p{7.2cm}@{}} \textbf{Title:} How can I know which is the source of an specific standard shared libraries?\\ \textbf{Body:} How can I get access to the source code of standard shared libraries?\end{tabular} \\ \midrule
\emph{\textbf{non-duplicate}} & \begin{tabular}[t]{@{}p{7.2cm}@{}} \textbf{Title:} Grafics on Thinkpad R50e\\ \textbf{Body:} After installing Ubuntu 12.04 LTS on a Thinkpad R50e, there is no graphics driver, seems to me. The display works, but in a magenta-red mode, video is yellowish, YouTube has no problem. In Details, there is no graphics driver.\end{tabular} & \begin{tabular}[t]{@{}p{7.2cm}@{}}\textbf{Title:} How to share files between Windows7(Guest) and Ubuntu 12.04(Host)?\\ \textbf{Body:} I searched on the internet but all issues have Ubuntu as the Guest. I have VMWare Workstation 8 wherein Windows 7 is installed.\end{tabular} \\ \bottomrule
\end{tabular}
\label{table 6}
\end{table*}

\textbf{RQ4:} \emph{\textbf{What is the impact of domain-specific word embeddings on the performance improvement of ASIM? }}

\textbf{Motivation.} In this work, we built a 15GB corpus based on Stack Overflow data dump and trained word embeddings specific to the domain of software engineering on this corpus. We want to know whether and to what extent the domain-specific word embeddings improve the results compared with the word embeddings in the general domain. The answer to this research question helps us understand the improvement of using our pre-trained word embeddings for this task.

\textbf{Approach.} We replaced the word embeddings used in the experiment with the GloVe word embeddings \footnote{\url{http://nlp.stanford.edu/data/glove.840B.300d.zip}} consistent with Shirani \emph{et al.} and experimented. The rest of the model remains unchanged and compared with the original model on the F1-score metric.

\textbf{Result.} Table \ref{table 5} presents the experiment result. After replacing the word embeddings with GloVe word embeddings, the ASIM's performance drops slightly, which proves the importance of domain-specific word embeddings.

\begin{framed}
\textbf{RQ4:} Compared with GloVe word embeddings, using the word embeddings specific to the software engineering domain can make ASIM achieve better performance. So, appropriate domain-specific word embeddings are also important in some tasks.
\end{framed}

\textbf{RQ5:} \emph{\textbf{How is the generalization ability of ASIM? }}

\textbf{Motivation.} The duplicate question detection in PCQA websites is also a task to study the semantic relatedness in software engineering. Compared with predicting question relatedness, the most significant difference is that this task requires boolean prediction to determine whether a pair of questions is duplicate. We hope to apply ASIM to a similar but different software engineering task to explore the generalization performance and its robustness.

\textbf{Approach.} For the experiment dataset, we choose the no-clue version of the AskUbuntu duplicate question detection dataset (hereafter, AskUbuntu dataset) constructed by Rodrigues \emph{et al.} \cite{rodrigues2017ways}. In the AskUbuntu dataset, 24K question pairs are used for training, 6K for testing, and 1K for validation. The two classes of the AskUbuntu dataset are balanced, thus with an equal number of duplicate and non-duplicate question pairs. We randomly select two question pair samples of each class and list them in Table \ref{table 6}. Since the dataset only contains questions but not answers, we concatenate the question's title and body as the model's input and modify the Prediction Layer's output to 2 classes. The other part of the model remains unchanged. We use the same set of hyper-parameters tuned in the Knowledge Unit dataset' validation set. The ASIM is trained for 30 epochs and compares the model's accuracy on the test set with the DOTBILSTM, SOFTSVM, and the baseline models in the paper \cite{rodrigues2017ways}. The results of the DOTBILSTM and SOFTSVM are from the paper \cite{amirreza2019question}, and the results of the other baseline models are from the paper \cite{rodrigues2017ways}.

\textbf{Result.} Table \ref{table 7} presents the results. ASIM significantly outperforms other models, including rule-based approaches (Jcrd \cite{wu2011efficient}), classifiers (SVM-bas \cite{bogdanova2015detecting}, SVM-adv \cite{rodrigues2017ways}, and SOFTSVM), and neural networks (CNN \cite{bogdanova2015detecting}, DNN \cite{afzal2016mayonlp}, DCNN \cite{rodrigues2017ways}, and DOTBILSTM), achieves an accuracy of 96.25\%. Moreover, DCNN is the best model in the paper \cite{rodrigues2017ways} on the AskUbuntu dataset, which combines the CNN and DNN, and the SOFTSVM is the state-of-the-art model on this dataset.

\begin{table}
\caption{Performance of different models in the AskUbuntu dataset.}
\centering
\begin{tabular}{ll}
\toprule
\textbf{Models}      & \textbf{Accuracy} \\
\midrule
Jcrd        & 0.7291 \\
SVM-bas     & 0.7025 \\
SVM-adv     & 0.7587 \\
CNN         & 0.7450 \\
DNN         & 0.7865 \\
DCNN & 0.7900     \\
DOTBILSTM   & 0.87     \\
SOFTSVM     & 0.90     \\
\midrule
ASIM   & \textbf{0.9625}     \\
\bottomrule
\end{tabular}
\label{table 7}
\end{table}

\begin{framed}
\textbf{RQ5:} ASIM can also achieve good performance in the duplicate question detection task of AskUbuntu, a different software engineering task. ASIM outperforms the SOTA model on this dataset and proves its generalization ability and robustness.
\end{framed}

\section{DISCUSSION}
In this section, we present the visualization of attention with examples to show the attention mechanism's capability in ASIM. Then, we discuss threats to our approach's validity.

\subsection{Attention Visualization}
\begin{figure}[h]
  \centering
  \subfigure[Attention visualization of the first question title pair.]{
  \includegraphics[width=0.9\linewidth]{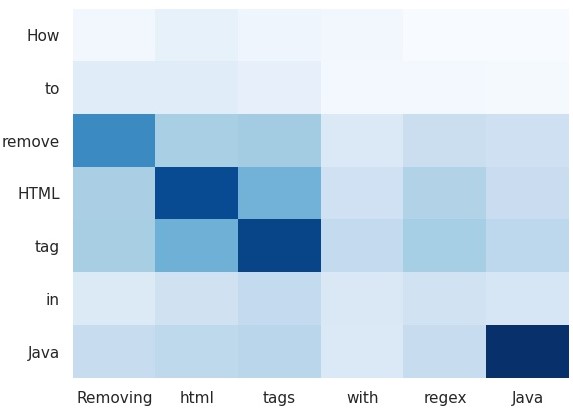}
  \label{case a}
  }
  \centering
  \subfigure[Attention visualization of the second question title pair.]{
  \includegraphics[width=0.9\linewidth]{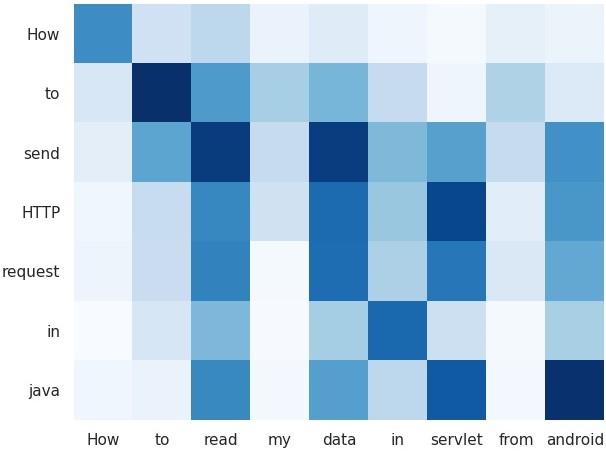}
  }
  \caption{Visualization of attention results for two examples.}
\label{fig:attention}
\end{figure}

We present a case study through attention visualization to investigate what ASIM learns in the Attention Layer. We pick titles of two duplicate question pairs from the Knowledge Unit dataset. The question titles in the first knowledge unit are "How to remove HTML tag in Java" and "Removing html tags with regex Java", and the question titles in the second knowledge unit pair are  "How to send HTTP request in java" and 'How to read my data in servlet from android'. We only present the question's title because the full knowledge unit's text sequence is too long. We visualize the word-by-word similarity based on the Inter-Attention matrix from \eqref{equation 7}. The attention results are shown in Figure \ref{fig:attention}, and a darker blue indicates a stronger similarity value in the attention matrix.

We can see that, through the attention mechanism, the important words are emphasized in both cases. Specifically, in case (a), because there are many common words between the two questions, the model can easily find these important words. Also, the model tends to align longer phrases together instead of individual words (e.g., the phrase "remove html tags"). In case (b), although there are fewer common words between the two sentences, the model can capture the semantic relatedness between words through learning and make the correct alignment. For example, the word "HTTP" is associated mostly with "servlet", and the word "java" is strongly associated with "android" in this case. Judging by the aligned words and phrases, ASIM can correctly classify the two labels as \emph{duplicate}.

\subsection{Threats to Validity}
In this part, we discuss the threats to construct, internal and external validity.

\textbf{Construct validity} relates to the correct identification of measures used in the measurement procedure\cite{wang2019detecting}. We use Precision, Recall, and Micro-F1 as evaluation metrics, which were also used in previous work to evaluate the model's effectiveness in predicting the question relatedness \cite{xu2016predicting,amirreza2019question,fu2017easy}.

\textbf{Internal validity} relates to errors in our code and the experiment bias. To reduce the code errors, we double-checked and comprehensively tested the code, but there may still be errors that we overlooked. Moreover, The experimental dataset has been used in previous work \cite{amirreza2019question}. We also randomly selected experimental data for inspection to ensure the accuracy of the labels.

\textbf{External validity} relates to the generalization performance of our study. To evaluate our approach's generalization performance, we also applied ASIM on the AskUbuntu duplicate question detection task, another semantic relatedness task in software engineering. However, it is unclear whether our approach can be applied to more related tasks. In future work, we will apply the model to other semantic relatedness tasks in software engineering to further evaluate its generalization performance.

\section{Conclusion and Future Work}
In this paper, we propose a deep learning model ASIM based on the attention mechanism to predict the relatedness between Stack Overflow questions, which can better help programmers obtain information and solve problems. Besides, we pre-train word embeddings specific to the domain of software engineering, based on the corpus collected from the Stack Overflow data dump. The experiment results show ASIM's effectiveness and consistency in predicting question relatedness, outperforming the two baseline models SOFTSVM and DOTBILSTM in previous work. Moreover, in the duplicate question detection task of AskUbuntu, ASIM can also achieve state-of-the-art performance, which proves its generalization ability. We will explore the application that predicts the relatedness between a newly posted question and an already answered question in future work.

\bibliographystyle{IEEEtran}
\bibliography{ref}

\end{document}